\documentclass[sigconf]{acmart}

\usepackage{booktabs} 
\usepackage{pdfpages}
\usepackage{graphicx}
\usepackage{subfig}

\setcopyright{rightsretained}

\acmDOI{10.475/123_4}

\acmISBN{123-4567-24-567/08/06}

\acmConference[ODD'18]{ACM SIGKDD 2018 Workshop
August 20th, London UK}
\acmYear{2018}
\copyrightyear{2018}

\acmArticle{4}
\acmPrice{15.00}

\begin{document}
\title{Anomaly Detection in the Presence of Missing Values}
\author{Thomas G. Dietterich}
\orcid{0000-0002-8223-8586}
\affiliation{%
  \institution{Oregon State University}
  \streetaddress{1148 Kelley Engineering Center}
  \city{Corvallis}
  \state{Oregon}
  \postcode{97331-5501}
}
\email{tgd@cs.orst.edu}
\author{Tadesse Zemicheal}
\affiliation{%
  \institution{Oregon State University}
  \streetaddress{1148 Kelley Engineering Center}
  \city{Corvallis}
  \state{Oregon}
  \postcode{97331-5501}
}
\email{zemichet@oregonstate.edu}

\begin{abstract}{
Standard methods for anomaly detection assume that all features are observed at both learning time and prediction time. Such methods cannot process data containing missing values. This paper studies five strategies for handling missing values in test queries: (a) mean imputation, (b) MAP imputation, (c) reduction (reduced-dimension anomaly detectors via feature bagging), (d) marginalization (for density estimators only), and (e) proportional distribution (for tree-based methods only). Our analysis suggests that MAP imputation and proportional distribution should give better results than mean imputation, reduction, and marginalization. These hypotheses are largely confirmed by experimental studies on synthetic data and on anomaly detection benchmark data sets using the Isolation Forest (IF), LODA, and EGMM anomaly detection algorithms. However, marginalization worked surprisingly well for EGMM, and there are exceptions where reduction works well on some benchmark problems. We recommend proportional distribution for IF, MAP imputation for LODA, and marginalization for EGMM.}
\end{abstract}

%
%




\maketitle
\section{Introduction}
In supervised learning, the problem of missing values in training and test data has been studied for many years \cite{quinlan2014c4,saar2007handling}. In the past decade, many general purpose algorithms have been developed for anomaly detection \cite{Emmott2015,Zimek2012,Chandola2009}. However, with few exceptions,  there are no published methods for handling missing values at either training time or prediction time for anomaly detection algorithms. 

This paper presents five methods for handling missing values in test queries. The paper defines the methods, analyzes them formally, and studies them experimentally.  The paper is organized as follows. First, we describe three of the leading general purpose anomaly detection algorithms, the Isolation Forest (IF) \cite{liu2008isolation}, LODA \cite{pevny2016loda}, and EGMM \cite{Emmott2015}. Recent benchmarking studies by \citet{Emmott2015} have shown that these are among the most effective algorithms in the literature.  Then we review existing research on general methods for handling missing values in machine learning and describe how these general methods are instantiated for IF, LODA, and EGMM. We then define and analyze the methods formally and develop a set of hypotheses. Finally, we test these hypotheses on synthetic data and benchmark data sets and summarize the results. From the analysis and experiments, we conclude that the method of MAP imputation works well across all three algorithms but that for IF, the method of proportional distribution is more efficient, and for EGMM, the method of marginalization gives better performance.

\section{Unsupervised Anomaly Detection Algorithms}
We study the standard setting of unsupervised anomaly detection. At training time, we are given a data set $\mathcal{D}_{\mathrm{train}} = \{X_1,\ldots, X_N\}$, where $X_i \in \mathbf{R}^d$. We will index the features of each $X_i$ by the subscript $j=1, \ldots, d$. The data comprise an unknown (and unlabeled) mixture of ``nominal'' and ``anomaly'' points. We assume that the anomalies make up only a small fraction of $\mathcal{D}_{\mathrm{train}}$. At test time, we are given a second data set $\mathcal{D}_{\mathrm{test}}$ that is also a mixture of nominal and anomaly points. An anomaly detection algorithm analyzes $\mathcal{D}_{\mathrm{train}}$ and constructs an anomaly detector $A: \mathbf{R}^d \mapsto \mathbf{R}$, which is a function that assigns a real-valued anomaly score $A(X_q)$ to a $d$-dimensional query vector $X_q$. The ideal anomaly detector assigns high scores to the anomalous points and low scores to the nominal points such that we can set a threshold that perfectly separates these two classes.  Anomaly detectors are typically evaluated according to the area under the ROC curve (AUC), which is equal to the probability that the anomaly detector will correctly rank a randomly-chosen anomaly point above a randomly-chosen nominal point.

The definition of an anomaly is domain-dependent, but our basic assumption is that the anomaly points are being generated by a process that is different from the process that generates the nominal points. Virtually all algorithms for unsupervised anomaly detection work by trying to identify outliers and then assuming that the outliers are the anomalies. We call this strategy ``anomaly detection by outlier detection''.

We now describe each of the three anomaly detection algorithms studied in this paper.

\subsection{Isolation Forest}

The Isolation Forest (IF) algorithm constructs an ensemble of isolation trees from the training data. Each isolation tree is a binary tree where each internal node $n$ consists of a threshold test $X_j \geq \theta_n$. IF constructs the isolation tree top-down starting with the entire training set $\mathcal{D}_{\mathrm{train}}$ at the root node. At each node $n$, IF selects a feature $j$ at random, computes the minimum $\underline{M}_j$ and maximum $\overline{M}_j$ values of the data $\mathcal{D}_n$ that arrive at this node, selects $\theta_n$ uniformly at random from the interval $(\underline{M}_j, \overline{M}_j)$, and then splits the data into two sets: $\mathcal{D}_{\mathrm{left}}$ of points $X_i$ for which $X_{i,j} \geq \theta_n$ and $\mathcal{D}_{\mathrm{right}}$ of points $X_i$ for which $X_{i,j} < \theta_n$. Splitting continues until every data point is isolated in its own leaf node. In the IF paper, the authors describe methods for early termination of the tree-growing process, which can give large speedups, but we do not employ those in this paper.

IF creates an ensemble of trees by building each tree from a subset of the training data. In the original IF paper, the authors grow an ensemble of 100 trees using samples of size 256 drawn uniformly without replacement from the training data. 

At test time, an anomaly score is assigned as follows. The query point $X_q$ is ``dropped'' down each tree $t=1,\ldots,T$ to compute an isolation depths $d_1, \ldots, d_T$. The anomaly score is computed by normalizing the average $\overline{d}$ of these isolation depths according to the formula $A(X)= \exp[-\frac{\overline{d}}{z}]$, where $z$ is a theoretically-determined expected isolation depth (computed from $N$). Hence, the smaller the value of $\overline{d}$, the larger the anomaly score. The computation of the isolation depth is somewhat subtle. Consider query point $X_q$ as it reaches node $n$ at depth $d$ in the tree. If $n$ is a leaf node, then the isolation depth is $d$. If $n$ is an internal node (with test feature $j$ and threshold $\theta_j$), then an early  termination test is performed. If $X_{q,j}<\underline{M}_j$ or $X_{q,j}>\overline{M}_j$, then the isolation depth is $d$. Otherwise, $X_q$ is routed to the left or right child node depending on the outcome of the test $X_{q,j} \geq \theta_j$.

\subsection{Lightweight Online Detector Of Anomalies (LODA)}

The LODA algorithm constructs an ensemble of $T$ one-dimensional histogram density estimators. Each density estimator, $p_t$,  $t=1,\ldots,T$ is constructed as follows. First, LODA constructs a projection vector $w_t$ by starting with the all-zero vector, selecting $k = \sqrt{d}$ features at random, and replacing those positions in $w_t$ with standard normal random variates. Each training example $X_i$ is then projected to the real line as $w_t^\top X_i$, and a fixed-width histogram density estimator $p_t$ is estimated via the method of \citet{Birge2006}. To compute the anomaly score of a query point $X_q$, LODA computes the average log density $f(X_q) = 1/T \sum_{t=1}^T -\log p_t(w_t^\top X_q)$. In our experiments, we set $T=100$. 

\subsection{Ensemble of Gaussian Mixture Models (EGMM)}

The EGMM algorithm \cite{Senator2013,Emmott2015} learns an ensemble of Gaussian mixture models (GMMs) by fitting 15 GMMs to bootstrap replicates of the training data for each number of mixture components $k=3, 4, 5$. This gives 45 GMMs. For each $k$, we compute the mean out-of-bag log likelihood and keep only those models corresponding to values of $k$ whose mean log likelihood was at least 85\% of the log likelihood for the best $k$. This model selection step discards values of $k$ whose GMMs that do not fit the data well. The anomaly score for each test query $x_q$ is computed as $1/L \sum_{\ell=1}^L -\log p_\ell(x_q)$, where $L$ is the number of fitted GMMs and $p_\ell(x)$ is the density assigned by GMM $\ell$ to $x$.

\section{Missing Values and Methods for Handling Them}

Rubin \cite{rubin2004multiple} introduced a taxonomy of different kinds of missingness. In this paper, we study feature values missing completely at random (MCAR). In this form of missingness, the process that causes features to be missing in $X$ is completely independent from all of the feature values in $X$. 

Let $P$ be the probability density function of the population from which the training and test data sets $\mathcal{D}_{\mathrm{train}}$ and $\mathcal{D}_{\mathrm{test}}$ are drawn.  Note that $P$ assigns non-zero density to both nominal points and anomaly points.  Let $A: \mathbf{R}^d \mapsto \mathbf{R}$ be an anomaly detector that assigns a real-valued anomaly score to each point $X\in \mathbf{R}^d$. 

Let $J = \{1,\ldots,d\}$ be a set indexing the features and $M\subset J$ be a set of feature values that are missing in a query instance $X_q$.  Let $X_q[M]$ denote the features of $X_q$ indexed by $M$ and $X_q[-M]$ be the features indexed by the complementary set $J \setminus M$. 

\begin{definition} 
The {\em ideal method} for handling MCAR missing values is to compute the expected value of the anomaly score conditioned on the observed features:
\[A(X[-M]) = \int_{X[M]} P(X[M]\;|\;X[-M]) A(X) dX[M].\]
\end{definition}

Evaluating this high-dimensional integral is generally infeasible. One alternative is to approximate the integral by the conditional MAP estimate:

\begin{definition} {\em MAP imputation} computes 
\[\hat{X}[M] := \arg \max_{X[M]} P(X[M]\;|\;X[-M])\]
and returns the anomaly score $A([X[-M],\hat{X}[M]])$.
\end{definition}

An even weaker form of imputation is to just use the unconditional means:
\begin{definition} {\em Mean imputation} estimates the mean value
\[\overline{X} := \int_{X} P(X) X dX\]
and then returns the anomaly score $A[X[-M],\overline{X}[M]]$. 
\end{definition}

An alternative to approximating the ideal method is to learn a separate anomaly detector for $x_q$ that considers only the features in $-M$. Following the work on Sarr-Tsechansky and Provost, we will call this the ``reduced'' method.

\begin{definition} The {\em reduced method} for handling missing values is to train a separate anomaly detector $A[-M]$ based on only the features in the set $-M$ and return the score $A[-M](X[-M]).$
\end{definition}

The reduced method can be directly implemented by EGMM by marginalizing out the missing features prior to computing the surprise: 
\[A[M](X[M]) = - \log \int_{X[M]} P([X[M], -X[M]]) d X[M].\]
This is easily done for each Gaussian component in the each GMM in EGMM. 

It is informative to contrast this with the definition of the ideal method. If we employed log surprise as our anomaly detector under the ideal method, the anomaly score would be
\[A(X[-M]) = \int_{X[M]} P(X[M]\;|\;X[-M]) (-\log P([X[M],X[-M]])) dX[M],\]
where the integral is outside the logarithm.

A disadvantage of the reduced method is that we must compute a separate anomaly detector $A[-M]$ for each pattern of non-missing values. While this is easy for EGMM, it is much more difficult for other anomaly detectors. One way to approximate this is to create an ensemble of lower-dimensional anomaly detectors as follows. Let $A[-M_1], \ldots,$ $A[-M_L]$ be a set of lower-dimensional anomaly detectors constructed using $L$ different feature subsets $-M_1,\ldots,-M_L$. A common way to do this is known as``feature bagging". Define each subset $-M_i$ by selecting a set features of size $k < d$ at random and then training an anomaly detector using only those features.  Then, given a query $X[-M]$, we can return the average anomaly score using all anomaly detectors in the set $U[M] = \{A[-M_i] : -M_i \subseteq -M\}$. 
\[A_{\mathrm{reduced}}(X[-M]) = \frac{1}{|U|} \sum_{A\in U[M]} A(X[-M]). \]
This is exactly the method that was introduced by Pevn{\`y} in LODA \cite{pevny2016loda}. It can be applied to the Isolation Forest by growing each isolation tree using a subset $-M_i$ of features. 

For tree-based classifiers, \citet{quinlan1989} introduced a method that we will call ``proportional distribution''. This method can be applied by Isolation Forest as follows. Consider computing the isolation depth in an isolation tree for a query point $X_q$. When the query reaches internal node $n$ with test $X_j \geq \theta_n$, if $X_{q,j}$ is missing, then the query is sent down both the left and right child trees, which recursively return isolation depth estimates $d_{\mathrm{left}}$ and $d_{\mathrm{right}}$. Node $n$ then returns the depth $1 + P_{\mathrm{left}} d_{\mathrm{left}} + P_{\mathrm{right}} d_{\mathrm{right}}$, where $P_{\mathrm{left}}$ is the proportion of training examples for which $X_j \geq \theta_n$ and $P_{\mathrm{right}} = 1 - P_{\mathrm{left}}$.  We can view proportional distribution as an approximation of the ideal method.

In this paper, we employ an approximation to MAP imputation based on the R ``mice'' package \cite{StefvanBuuren2011} as reimplemented in the python ``fancyimpute'' package. The mice imputation procedure works as follows. First, it fills in each missing value via mean imputation. Then it makes repeated Gibbs sampling passes through the test data set. In each pass and for each feature $j$,  mice fits a Bayesian linear ridge regression model to predict $X[j]$ as a function of all of the remaining variables.  We set the ridge regularization parameter to 0.01. Let $\beta[j]$ denote the fitted regression parameters.  To impute each missing value for feature $j$, mice samples a value from the posterior predictive distribution: $X_q[j] \sim P(X[j] | X_q[-j], \beta)$. We perform 110 passes and discard the first 10 (the ``burn-in period''). The final imputation for each feature is the mean of the 100 values imputed during the Gibbs sampling.

\subsection{Analysis of Missing Values Methods}\label{sec:analysis}

Let us consider the situations under which we expect the various missing values methods to work well.  What do we mean by ``work well''? Let $0 \leq \rho < 1$ be the proportion of feature values that are missing in a test query.  We expect that as $\rho$ increases, the performance of the anomaly detection algorithms will decay, even when we apply missing values methods. We will say that a method works well if its performance decays slowly as $\rho$ increases. 

Imputation is only possible if there are correlations among the feature values in the data. Hence, if all of the feature values are independent, we would not expect any of the imputation methods to work well---they should all give equivalent performance.

\begin{figure}
\centering
\includegraphics[width=\columnwidth]{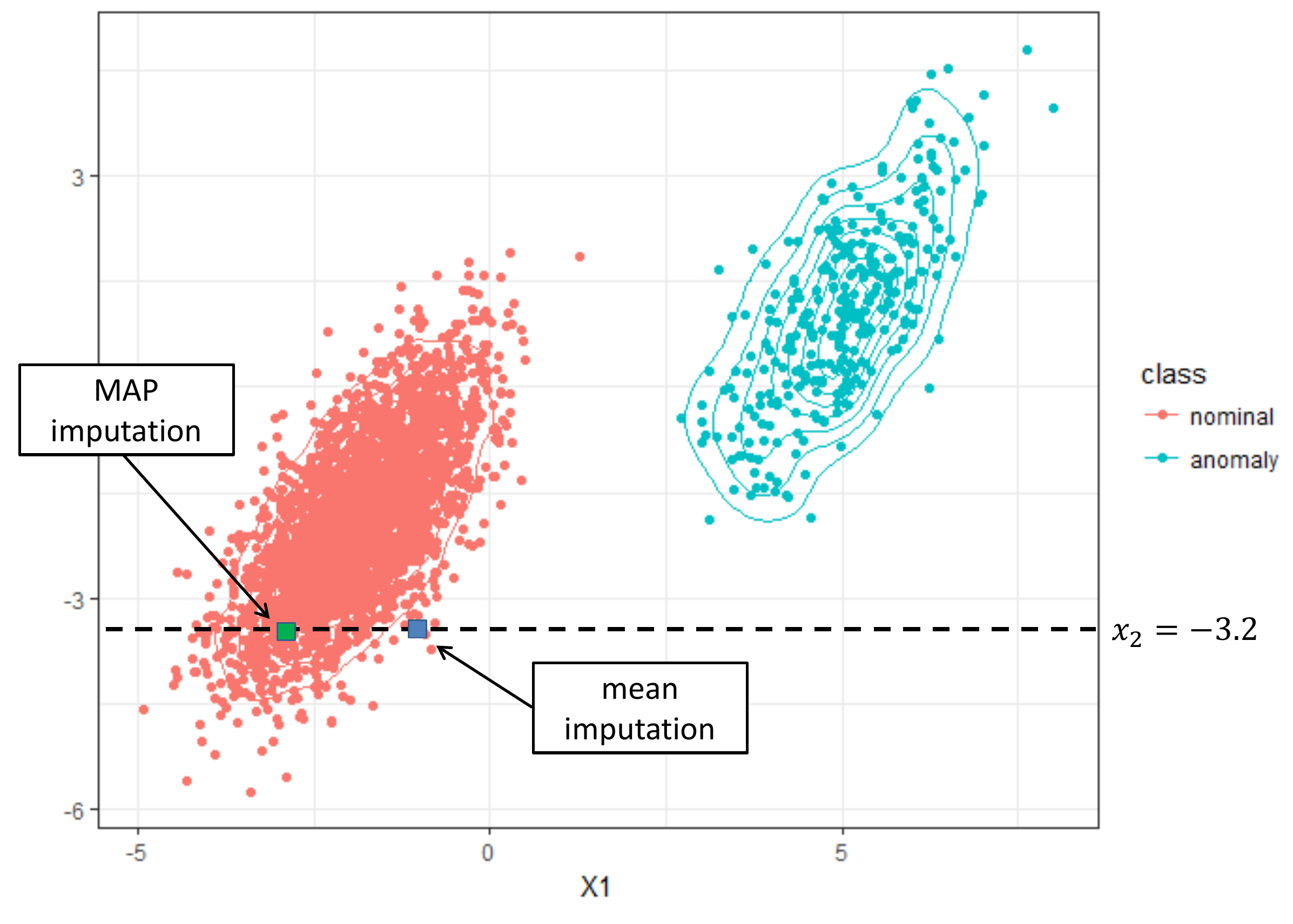}
\caption{A two-dimensional illustration of correlated feature values. The MAP imputation is the green square; the mean imputation is the blue square.}
\label{fig:mean-map}
\end{figure}

Now consider problems with feature correlations. When will MAP imputation work better than mean imputation? Precisely when the MAP estimate is different from the mean estimate. Figure~\ref{fig:mean-map} shows a simple example of a data set with 90\% nominal (red) and 10\% anomaly (turquoise) points. The two features are strongly correlated so that if we observe the query $X_q = (NA, -3.2)$, we can impute the missing value $X_1$. MAP imputation finds the highest density point along the line $X_2 = -3.2$, which is at $X_1=-2.31$, whereas mean imputation uses the overall mean value of $X_1$ which is $X_1=-1.20$. Because the mean-imputed value $(-1.20,-3.2)$ lies in a low-density region, the anomaly detectors will assign it a high anomaly score, whereas the MAP-imputed value of $(-2.31,-3.2)$ lies in a higher-density region and will be assigned a lower anomaly score.

\begin{figure}
\centering
\includegraphics[width=\columnwidth]{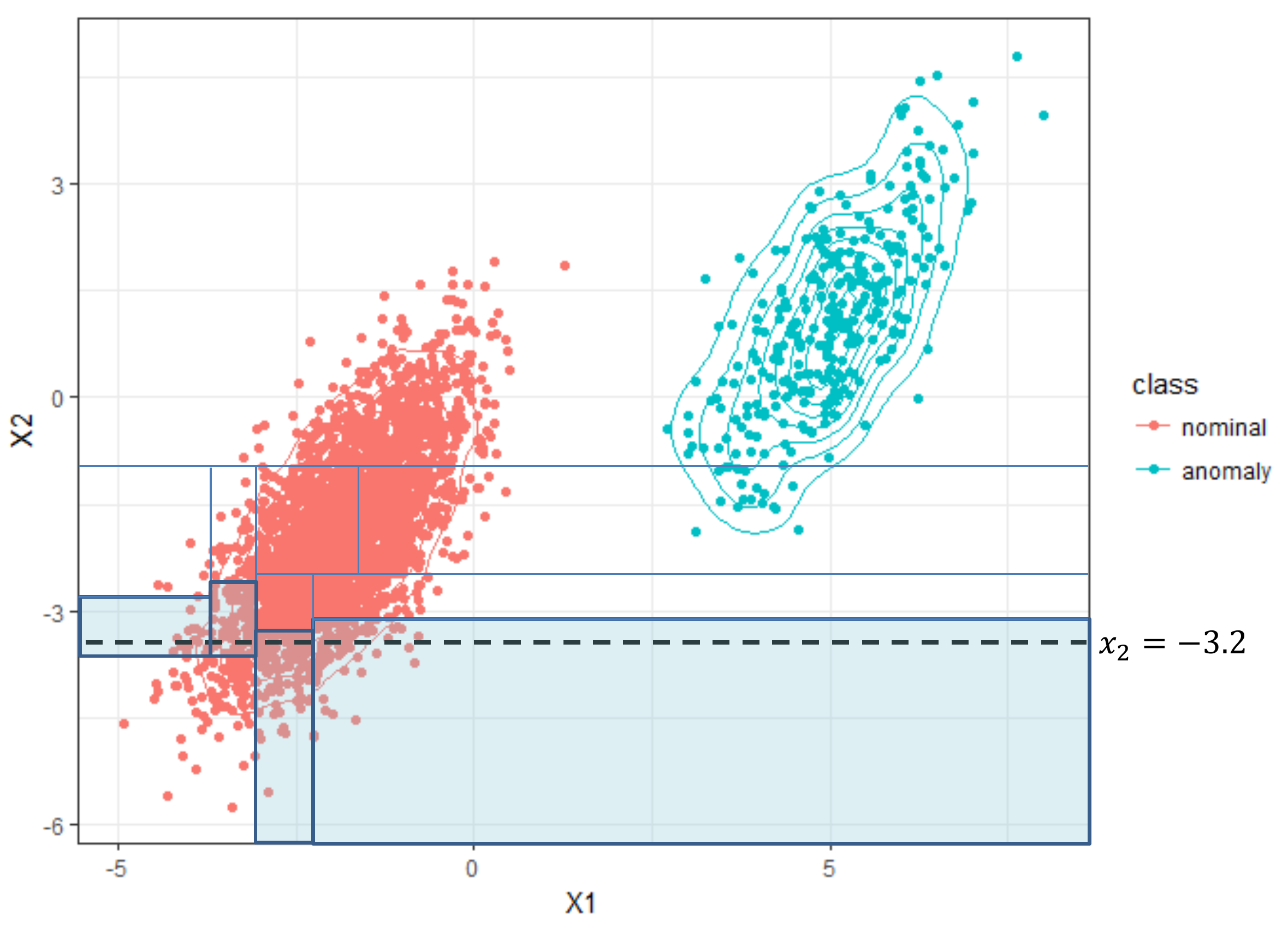}
\caption{Illustration of proportional distribution. Thin vertical and horizontal lines correspond to thresholds $\theta_j$ at various internal nodes in an isolation tree. Shaded regions indicate the subtrees whose isolation depths will be weighted and summed by proportional distribution.}
\label{fig:pd}
\end{figure}

How does proportional distribution behave? Figure \ref{fig:pd} shows what it does on this same example data set. Each isolation tree divides the feature space into axis-parallel rectangles recursively. In the figure, we only show the relevant splitting thresholds to depth 5 for a single tree. Each of the four shaded regions corresponds to a subtree rooted at depth 5, and the final isolation depth will be the weighted average of the isolation depths in these four regions. We can see that this is a crude approximation of integrating the isolation depth along the dashed line, which is what the ideal method would do. We can expect proportional distribution to work better than MAP imputation in cases where the conditional distribution of a missing value is multi-modal. MAP imputation will choose one mode; mice might average values from multiple modes and end up with an imputed value in the middle of empty space. Proportional distribution will integrate (approximately) across all modes.

\begin{figure}
\centering
\includegraphics[width=\columnwidth]{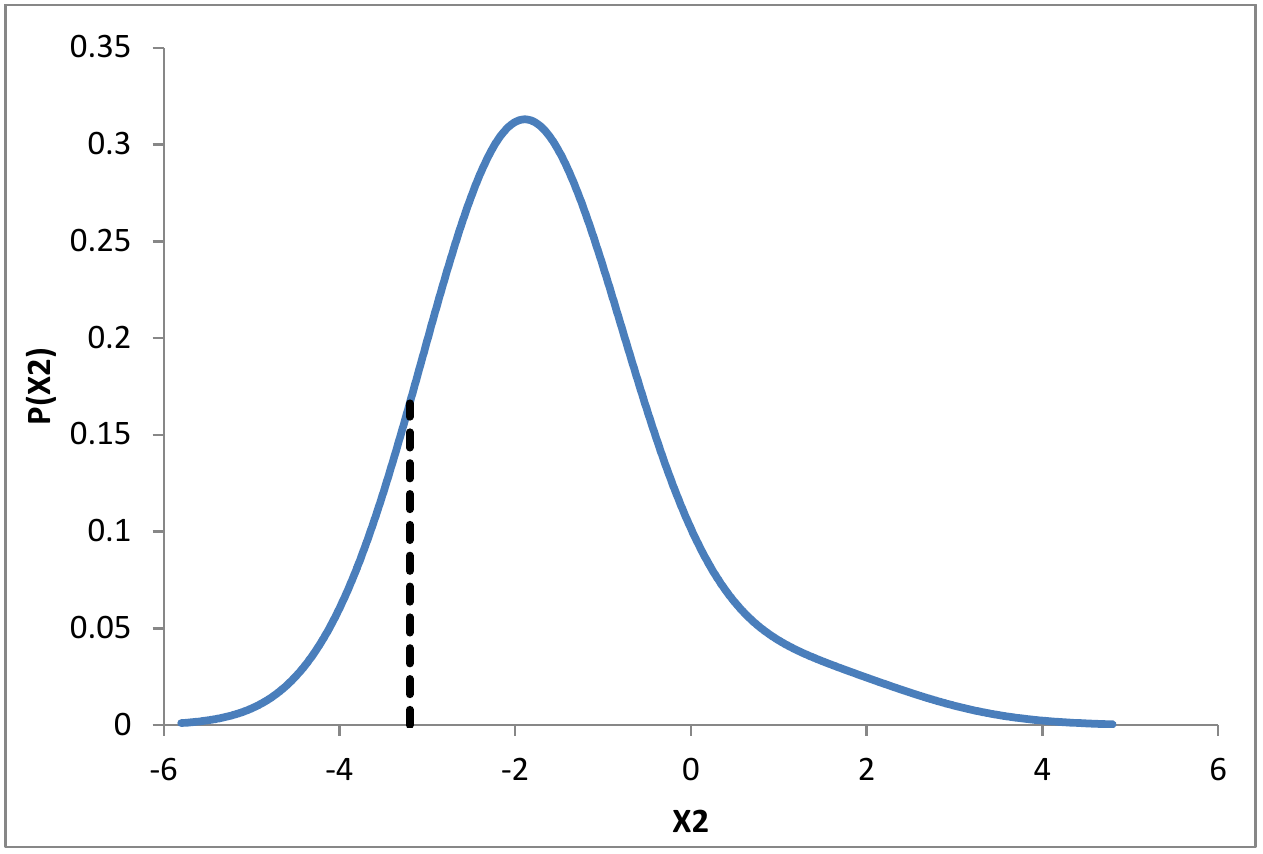}
\caption{Illustration of EGMM behavior. $X_1$ is marginalized away to obtain $P(X_2)$ and then the density at $X_2=-3.2$ gives the anomaly score.}
\label{fig:egmm}
\end{figure}

Finally, Figure~\ref{fig:egmm} shows how the marginalization method works. It plots the density $P(X_2)=\int_{X_1} P(X_1,X_2) dX_1$. The value at $P(X_2=-3.2)$ is fairly large, so EGMM will correctly assign a low anomaly score to this example.  

\section{Previous Studies of Missing Values in Supervised Learning}

\citet{saar2007handling} provide an excellent review of methods for handling missing values in supervised learning. They studied mean imputation, proportional distribution, and reduced classifiers. Sarr-Tsechansky and Provost found that proportional distribution did well in data sets where it was difficult to impute missing values correctly. But in data sets where imputation was possible---typically because of correlated information across multiple features---imputation generally worked better. They concluded that decision tree algorithms fail to include all redundant features when growing the tree and, consequently, proportional distribution cannot recover well from missing values.  Experiments by Quinlan \cite{quinlan1989,quinlan1987decision} also suggested that proportion distribution works best if the missing also missing at training time so that the decision tree is forced to learn to recover from missing values by testing additional features. Randomized ensembles, such as random forests and isolation forests, are less likely to suffer from this problem.

Hao \cite{hao2009} found that proportional distribution also works well for decision tree ensembles trained via gradient boosting. His studies focused on missing values in sequence-to-sequence mapping using conditional random fields.

The method that performed best in the studies by Sarr-Tsechansky and Provost was the ``reduced'' approach of training a separate classifier for each pattern of missing values. The main drawback is the cost in computation time and memory.

\section{Experimental Study}
To assess these methods for handling missing values, we conducted two studies: a study on synthetic data sets and a study on several UCI data sets \cite{Lichman:2013}.  We implemented our own versions of IF, LODA, and EGMM to support the missing value methods described above. 

\subsection{Synthetic data}
The goal of our synthetic data experiments is to evaluate the analysis of Section~\ref{sec:analysis}. We designed four synthetic data sets, each containing 3000 examples in 8-dimensional Euclidean space. In each data set, 10\% of the points are anomalies and the remaining 90\% are nominals. We describe each data set in turn along with the hypotheses that it is intended to test. Each experiment is repeated on 20 replications of each data set.

{\bf Uncorrelated:} In this data set, all features are independent. Nominals are sampled from a standard normal $\mathcal{N}(0,I)$ and anomalies are sampled from $\mathcal{N}(3,I)$. Our analysis predicts that mean imputation, MAP imputation, and proportional distribution will give very similar results while the reduced method may suffer from using lower dimensional projections of the data.

{\bf Noise:} For the second data set, we added 5 noise features to the samples in the uncorrelated data set. The noise features are sampled from $\mathrm{Unif}(-1, +1)$. We hypothesize that the reduced method may do better on this, because the additional noise features might confuse the imputation methods by introducing added variance.

{\bf Correlated:} In this data set, points are generated from ``correlated'' multivariate distributions. Details of the data generation process are given in the appendix.  Our analysis predicts that MAP imputation and proportional distribution will work better on correlated data than mean imputation. We expect the reduced method to perform worse because of its low-dimensional projections.

{\bf Mixture:} The data points are sampled from a Gaussian mixture model (GMM; see appendix for details) consisting of three clusters of nominal points positioned at the vertices of a triangle in 8-d space. The anomaly points are generated from a single Gaussian positioned near the center of that triangle. Our analysis suggests that proportional distribution may perform better on mixtures than either MAP imputation or mean imputation. 

For each data set and each anomaly detection algorithm, we performed the following experiment. First, we applied the anomaly detection algorithm to the data set to obtain a trained anomaly detector. Then we created nine test data sets by inserting fraction $\rho$ of missing values in each data point in the data set, for $\rho=0, 0.1, \ldots, 0.8$.  If $\rho \times d$ was not an integer, then we inserted missing values such that on average, fraction $\rho$ of the values were missing in each query example. For example, suppose $\rho=0.3$, $d=8$, and there are 1000 instances in the data set. To achieve an average of 2.4 missing values per instance, we insert 2 missing values in 600 of the instances and 3 missing values into the remaining 400 instances (chosen uniformly at random without replacement). 

\begin{figure}
\centering
\includegraphics[width=\columnwidth]{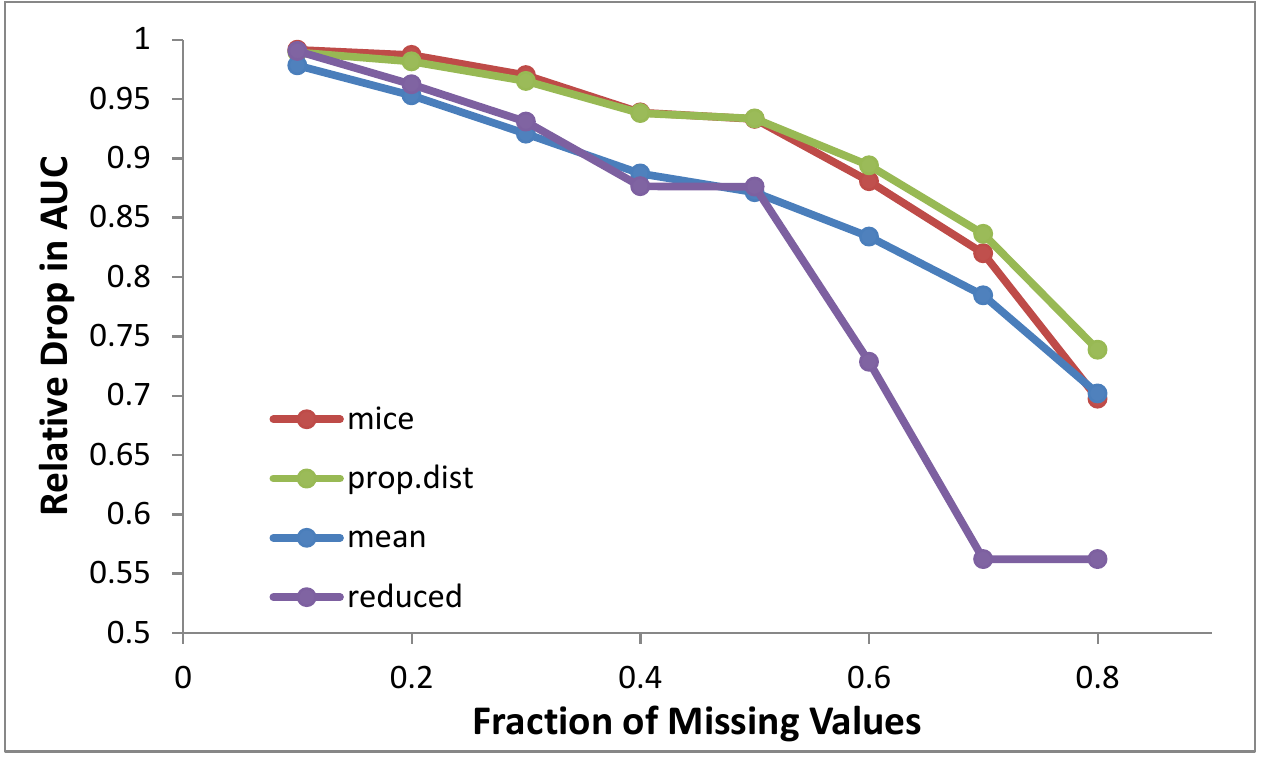}
\caption{Mean relative AUC of IF as a function of the fraction of missing values, $\rho$, averaged across the four synthetic data set configurations.}
\label{fig:synthetic-mean-decay-iforest}
\end{figure}

We then measured the AUC of the anomaly detector with each of its missing values methods on each test data set. For IF, the methods are mean imputation, mice imputation, proportional distribution, and reduced (using feature bagging with $\sqrt{d}$ features in each isolation tree). For LODA, the methods are mean imputation, mice imputation, and reduced. For EGMM, the methods are mean imputation, mice imputation, and marginalization. The AUCs of the 20 replicates of each condition (e.g., mixture + EGMM + marginalization) are averaged to obtain a final AUC.  We produce two summaries of this data: decay plots and 50\% missingness plots. Figure~\ref{fig:synthetic-mean-decay-iforest} shows how the AUC decays as $\rho$ increases. To normalize across the four synthetic data sets, we compute the relative AUC, which is equal to the AUC of each method divided by the AUC of the anomaly detection algorithm when $\rho=0$.  These relative AUC values are then averaged across the four synthetic data sets (i.e., over 20 replications of each synthetic configuration for a total of AUC 80 values). We observe that for IF, mice and proportional distribution give much better results than mean imputation and the reduced method. Note that the reduced method fails completely for $\rho=0.7$ and $\rho=0.8$, because there are fewer than $\sqrt{d}$ non-missing features, so none of the low-dimensional isolation trees can be evaluated.  Similar plots for LODA and EGMM are given in the appendix.

\begin{figure}
\centering
\includegraphics[width=\columnwidth]{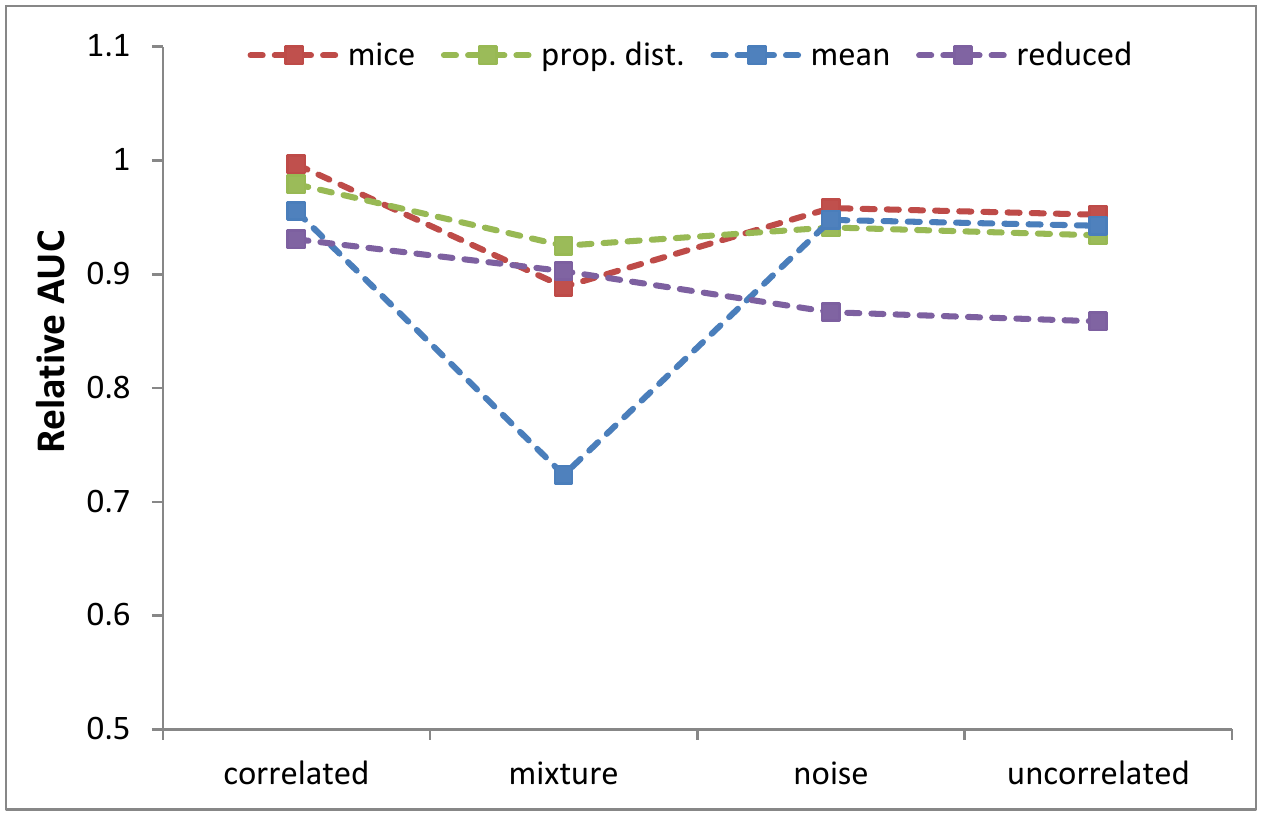}
\caption{Relative AUC of IF on the synthetic data sets for $\rho=0.5$, in ascending order by mice imputation.}
\label{fig:rho=0.5-iforest-synthetic}
\end{figure}

The second summary, shown in Figure~\ref{fig:rho=0.5-iforest-synthetic},  considers only $\rho=0.5$ and plots the relative AUC for each of the four synthetic configurations. This allows us to test the predictions arising from on our analysis. As expected, on the uncorrelated data, all methods give very similar performance, except that the reduced method is worse than the three imputation methods.  Adding noise slightly improves the performance of all methods, although this is not statistically significant. On correlated data, mice and proportional distribution perform well, but mean imputation begins to suffer somewhat. Finally, on the mixture data, we observe that proportional distribution is the best, as we predicted from our analysis. Mean imputation gives very bad results (because the mean often lies far from all of the data points), and the reduced method performs quite well. 

The experiments confirm our hypotheses with the exception that adding noise did not have any significant effect.

\subsection{UCI Benchmark Data Sets}
The goal of our second study is to see how well the missingness methods work on realistic data. We worked with the anomaly detection benchmarks constructed by Andrew Emmott \cite{Emmott2015} based on 13 supervised learning data sets from the UCI repository\cite{Lichman:2013}. Emmott refers to these 13 data sets as ``mother sets''. He defines a mother set by selecting some of the classes to be anomalies and the rest of the classes to be nominal data points. He has created thousands of benchmark data sets by systematically varying the proportion of anomaly points, the difficulty of the anomalies, the degree to which the anomaly points are clustered, and the level of irrelevant features. 

For our study, we chose 11 benchmark data sets from each mother set benchmark collection as follows: First, we ranked all benchmark data sets according the AUC scores computed using IF. We then chose the data sets ranked $290^{th}$ - $300^{th}$, where lower ranks are easier.  This ensures that our data sets are of medium difficulty, so that when we inject missing values, the effect will not be overshadowed by the difficulty (or easiness) of the benchmark data.  Table~\ref{table:data sets} summarizes the benchmark data sets. ``Mother Set type'' indicates the problem type of the original UCI mother set, $N$ indicates sample size (this can vary across the 11 benchmarks within each mother set), $d$ is the number of features, and ``\% anomalies'' gives the relative frequency of anomalies across the 11 benchmarks. 

\begin{table}
\caption{Description of the mother data sets.}
\begin{center}
	\resizebox{\columnwidth}{!}{%
		\begin{tabular}{cccccc} 
			\hline
			Name & Mother Set Type & $N$ & $d$ & (\%) anomalies\\
			\hline
			Abalone & regression & 1906 & 7 & (0.1\%, 0.5\%, 1\%, 10\%) \\ 
			Fault & binary & (277 - 1147) & 27 & (0.5\%, 1\%, 5\%, 16\%, 20\%) \\
            Concrete & regression & (399 - 422) & 8 & (0.6\%, 1\%, 5\%, 10\%) \\ 
			Image segment & multi class & (1190- 1320) & 18 & (0.5\%, 1\%,5\%,10\%) \\ 
			Landsat & multi class& (4593 - 4833) & 36 & (0.1\%, 0.5\%, 1\%,5\%,19\%) \\ 
			Letter recognition & multi class & 6000 & 16 & (1\%, 5\%, 10\%) \\ 
			Magic gamma Telescope & binary &  6000 & 10 & (0.1\%, 0.5\%, 1\%, 5\%)\\
			Page Blocks & multi class & 4600 & 55 &  (0.1\%, 0.5\%, 1\%, 5\%, 10\%) \\
			Shuttle & multi class & 6000 & 9 &  (0.1\%, 0.5\%, 1\%, 5\%, 10\%) \\ 
			Skin & binary & 6000 & 3 &  (0.1\%, 0.5\%, 1\%, 5\%, 10\%) \\ 
			Spambase & binary & (2512 - 2788) & 57 & (0.1\%, 0.5\%, 1\%, 5\%, 10\%) \\ 
			Wave & multi class & 3343 & 21 & (0.1\%, 0.5\%, 1\%, 5\%, 10\%)\\ 
            Wine & regression & (3720 - 4113) & 11 & (0.1\%, 0.5\%, 1\%, 5\%, 10\%) \\ 
            \hline
		\end{tabular}
	}
\end{center}
\label{table:data sets}
\end{table}

\begin{figure}
\centering
\includegraphics[width=\columnwidth]{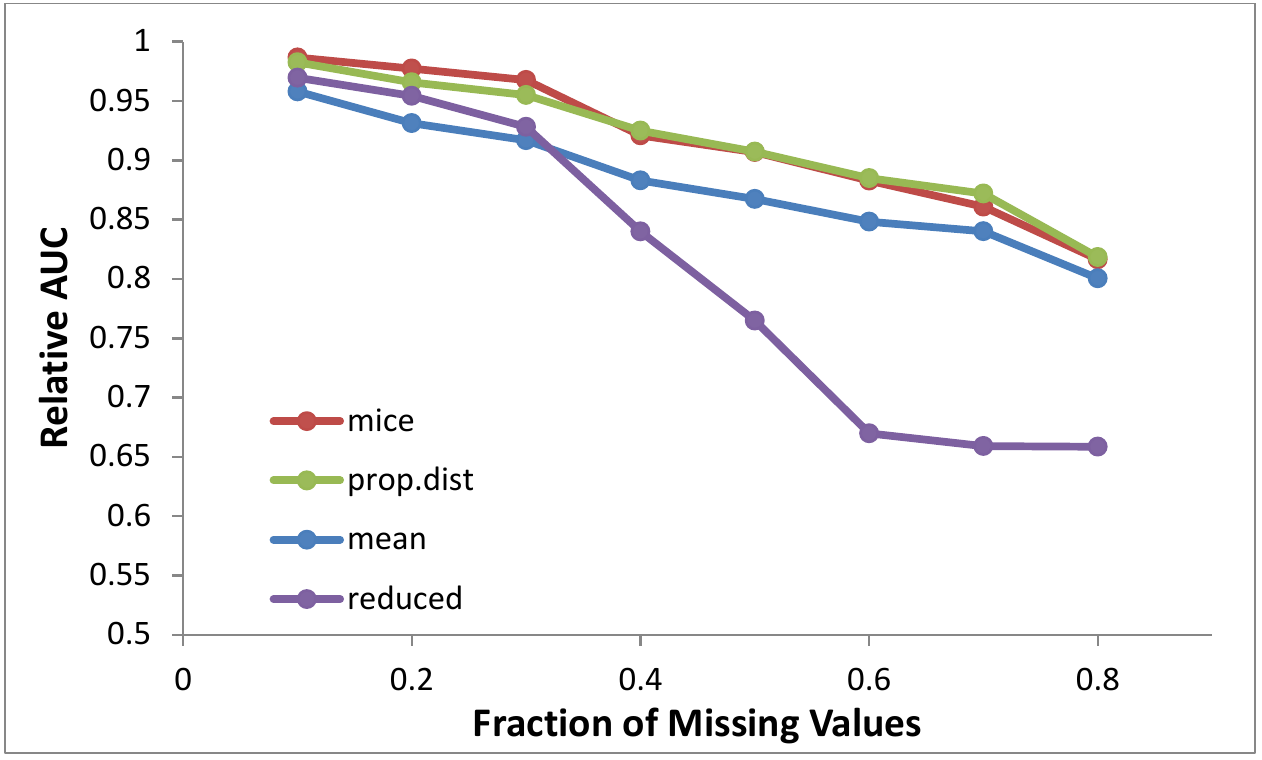}
\caption{Mean relative AUC of IF as a function of the fraction of missing values, $\rho$, averaged across 13 mother sets.}
\label{fig:mean-decay-iforest}
\end{figure}

\begin{figure}
\centering
\includegraphics[width=\columnwidth]{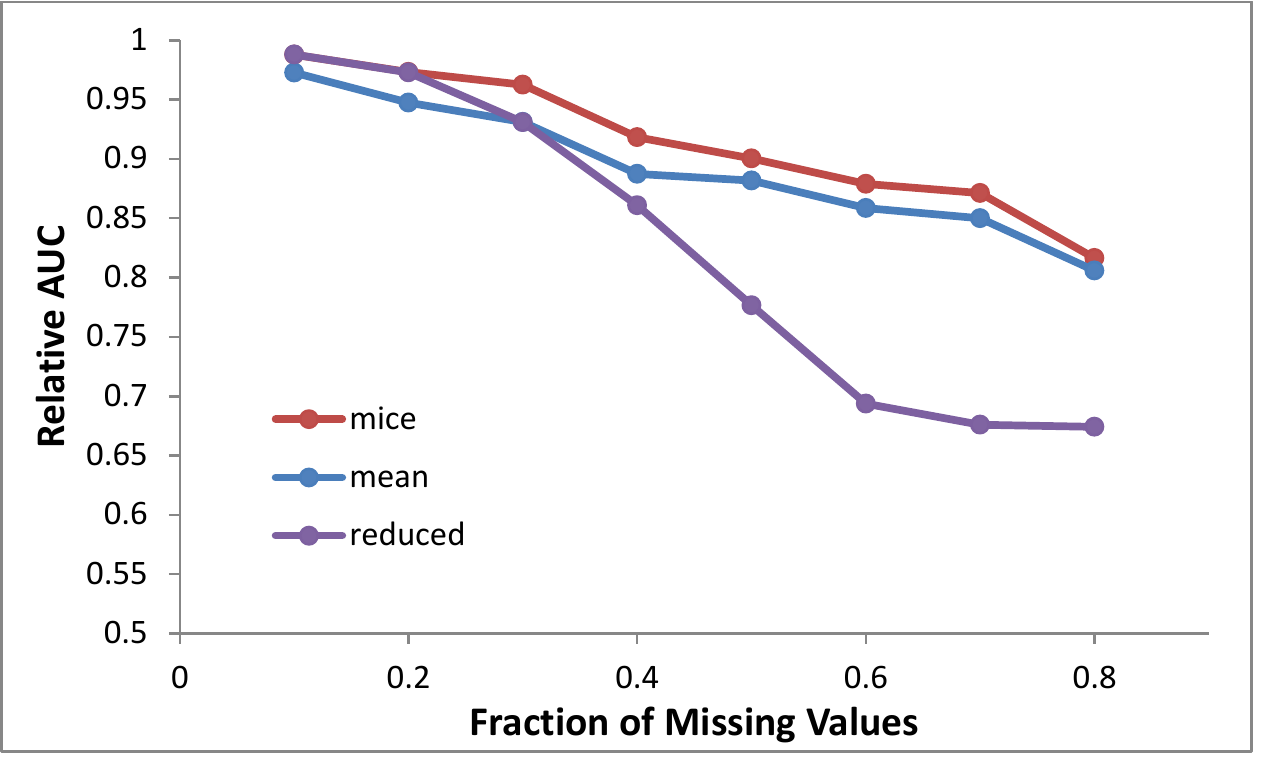}
\caption{Mean relative AUC of LODA as a function of the fraction of missing values, $\rho$, averaged across 13 mother sets.}
\label{fig:mean-decay-loda}
\end{figure}

\begin{figure}
\centering
\includegraphics[width=\columnwidth]{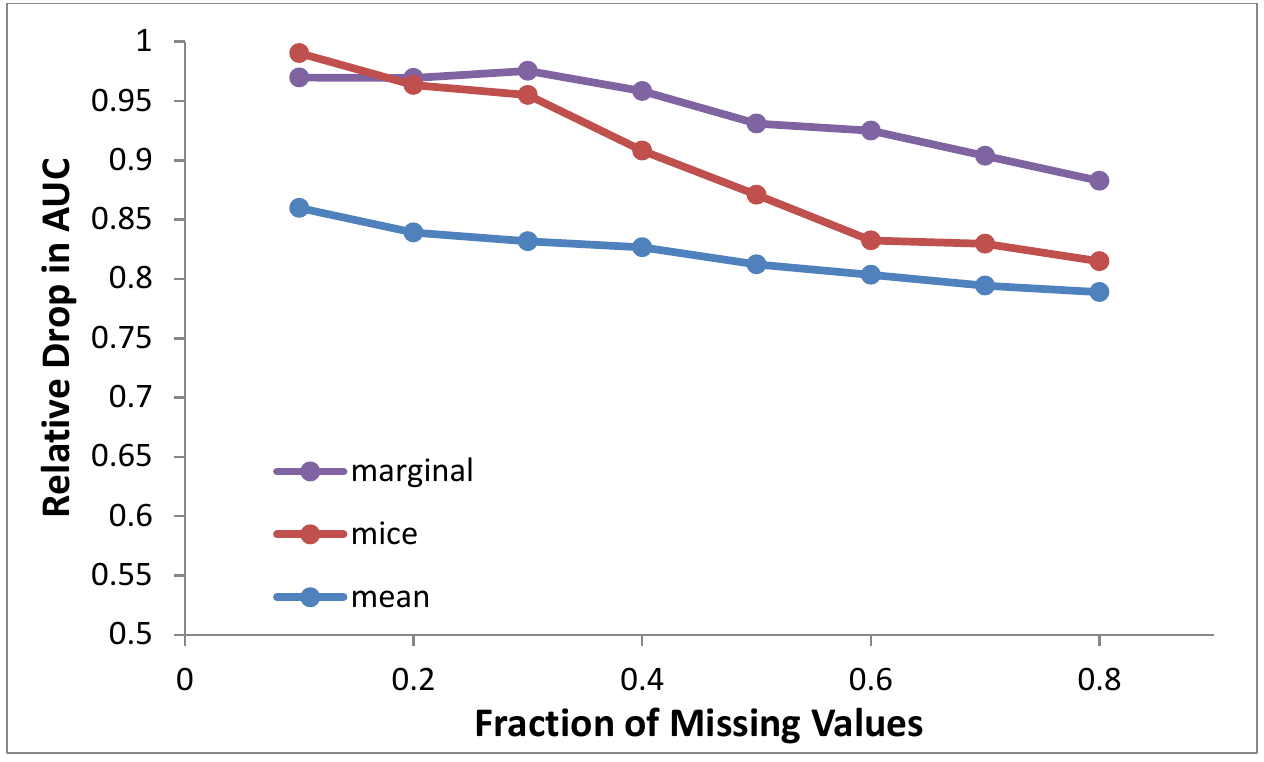}
\caption{Mean relative AUC of EGMM as a function of the fraction of missing values, $\rho$, averaged across 12 mother sets.}
\label{fig:mean-decay-egmm}
\end{figure}

\begin{figure}
\centering
\includegraphics[width=\columnwidth]{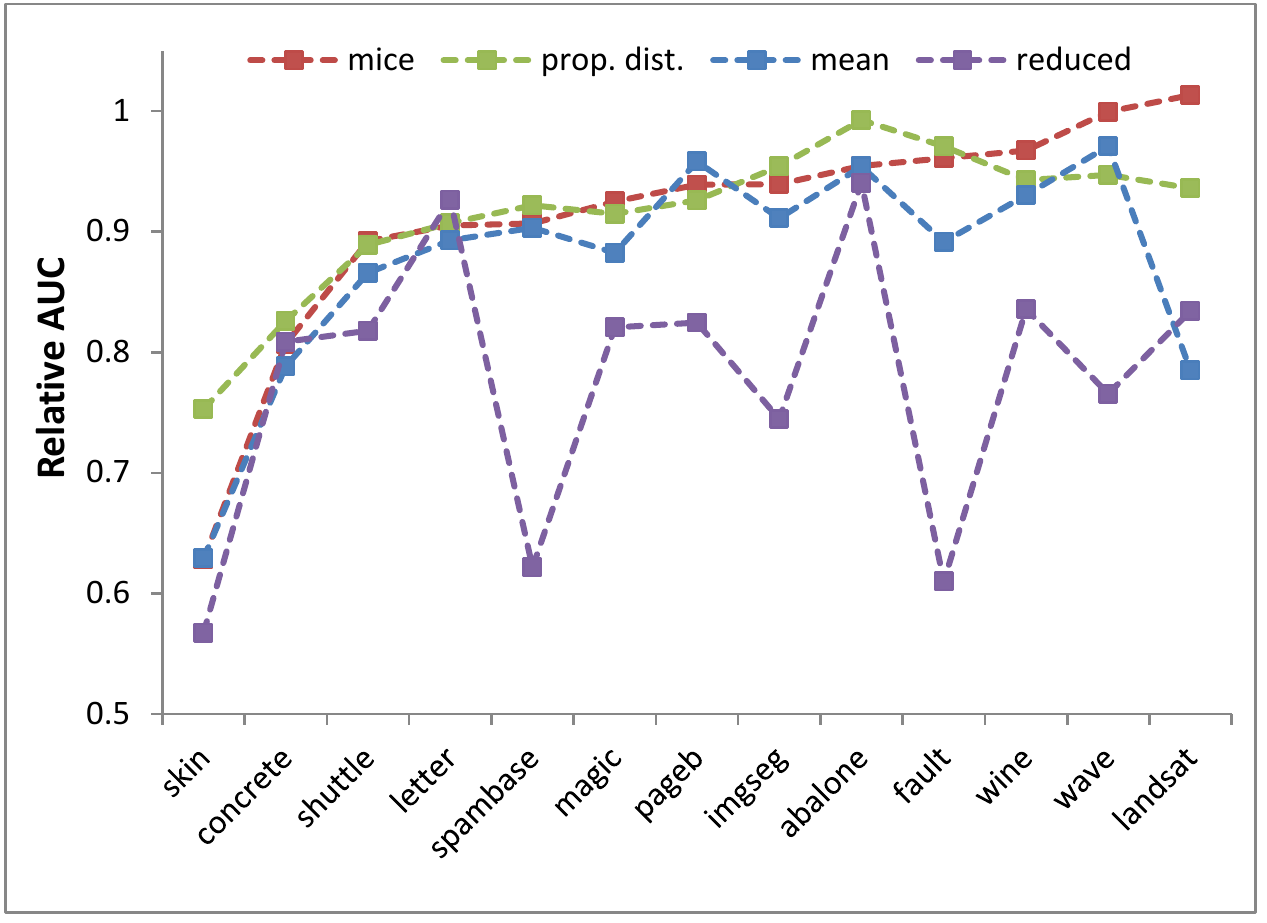}
\caption{Relative AUC of IF for $\rho=0.5$, in ascending order by mice imputation}
\label{fig:rho=0.5-iforest}
\end{figure}

\begin{figure}
\centering
\includegraphics[width=\columnwidth]{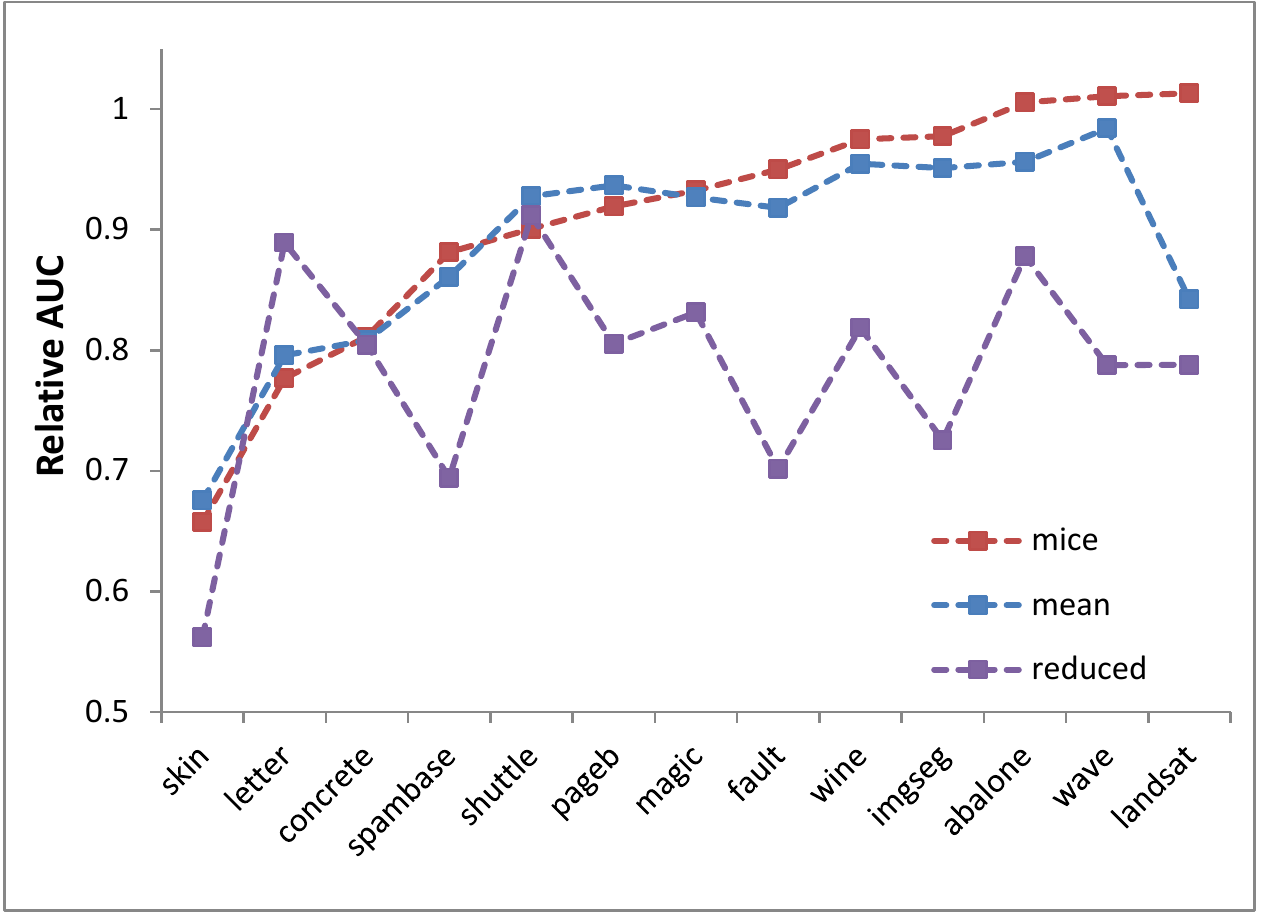}
\caption{Relative AUC of LODA for $\rho=0.5$, in ascending order by mice imputation}
\label{fig:rho=0.5-loda}
\end{figure}

\begin{figure}
\centering
\includegraphics[width=\columnwidth]{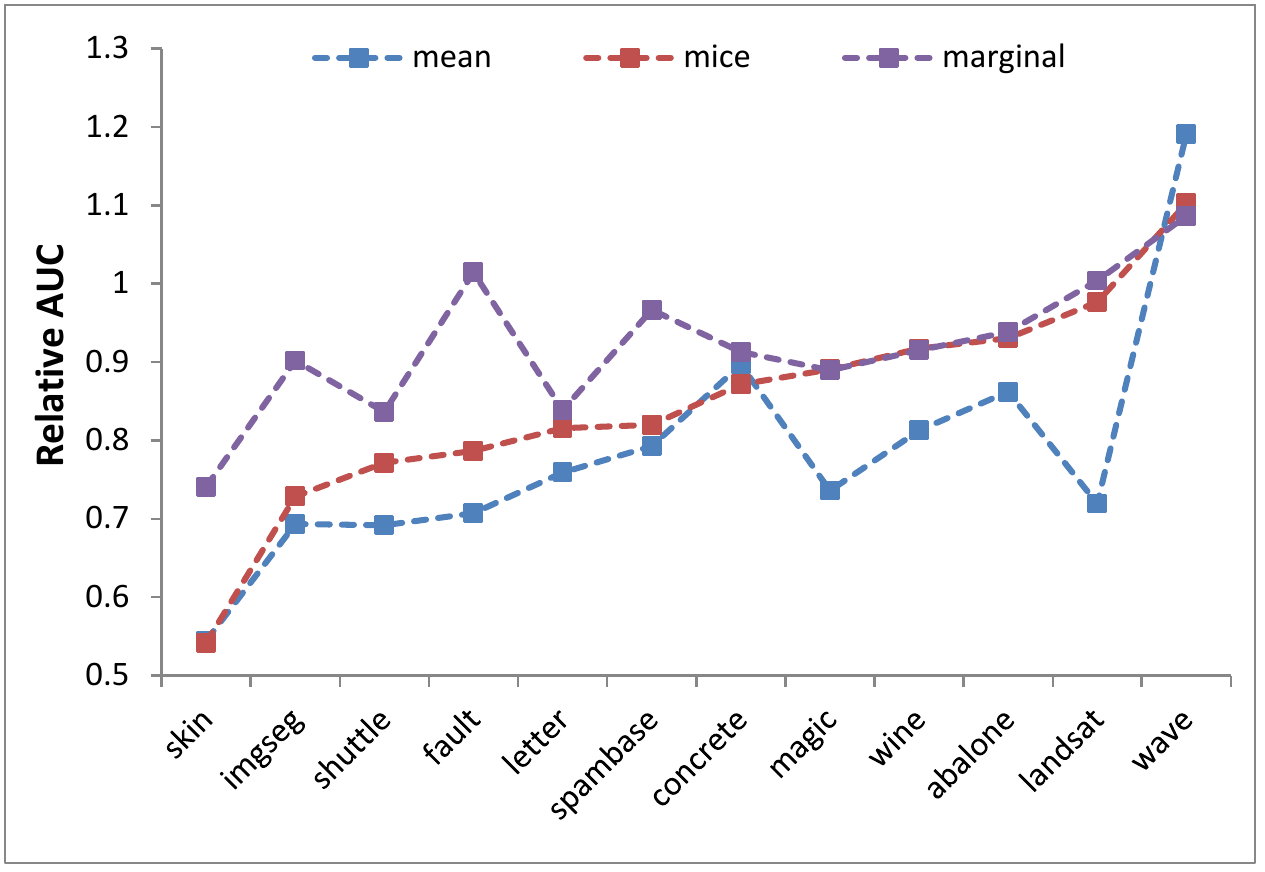}
\caption{Relative AUC of EGMM for $\rho=0.5$, in ascending order by mice imputation}
\label{fig:rho=0.5-egmm}
\end{figure}

Figures~\ref{fig:mean-decay-iforest}, \ref{fig:mean-decay-loda}, and \ref{fig:mean-decay-egmm} show how the AUC decays as $\rho$ increases. Figure~\ref{fig:rho=0.5-iforest}, \ref{fig:rho=0.5-loda}, and \ref{fig:rho=0.5-egmm} compare the relative AUC of the various missing values methods for $\rho=0.5$ across the different mother sets. 

The most striking result is that the reduced method performs very poorly for IF and LODA. This was expected for high values of $\rho$, because when the number of non-missing values falls below the number of features ($\sqrt{d}$) in the reduced anomaly detectors, the AUC falls to 0.5. But even at $\rho=0.1$, the reduced method is usually inferior to the other methods. There are occasional exceptions such as the letter recognition mother set for IF and LODA. A possible explanation for the poor performance of the reduced method is that each reduced anomaly detector has only $\sqrt{d}$ features, and therefore it is not able to assess the full normality/anomalousness of the test queries. 

Given the theoretical similarity between the reduced method and the marginal method for EGMM, it is surprising how well marginalization worked for EGMM. On average, for $\rho \geq 0.4$, the marginal method gave the best performance. Figure~\ref{fig:rho=0.5-egmm} shows that there are six mother sets for which the marginal method is best and only one (wave) where it is clearly inferior. Although the marginal method is not computing theoretically-correct anomaly scores, it is computing a very effective scores. Note that the marginalization method would require further refinement to work in practice. This is because it is not correct to compare a marginal distribution $p_1$ over $d_1$ dimensions to a distribution $p_2$ over $d_2 \neq d_1$ dimensions. All things being equal, if $d_1 < d_2$, then $p_1(x) \geq p_2(x)$, because $p_1$ disperses the probability density over a lower-dimensional space. In our experiments this was not a problem, because all of our test queries had the same (or nearly the same) number of missing features. In practice, it would be important to compute the tail probability $\int_{X} p(X) \mathbb{I}[p(X) \leq p(X_q)]dX$ of all points $X$ whose density is less than the density of the query point $p(X_q)$. Unlike densities, it is safe to compare probabilities. 

Now let's consider the three imputation methods: mice, mean, and proportional distribution. A very robust finding is that mean imputation performs worse than mice or proportional distribution. This was expected based on our analysis. Another observation is that mice and proportional distribution perform nearly identically, although there are some exceptions: Figure~\ref{fig:rho=0.5-iforest} shows that proportional distribution is clearly superior to mice on the skin and abalone mother sets and inferior on wine, wave, and landsat. Further study is needed to understand whether this is due to the cluster structure of skin and abalone as predicted by our analysis.

\section{Conclusions}
Based on our analysis and experiments, we make the following recommendations. First, implementations of Isolation Forest, LODA, and EGMM should always include algorithms for handling missing values. These algorithms never hurt performance, and when missing values are present, they can significantly improve performance. Second, the proportional distribution method works well for Isolation Forest. As it is more efficient than mice, it is our first choice method.  Third, mice imputation gave the best results for LODA---much better than Pevn{\`y}'s reduced method. We therefore recommend it as the first choice for LODA. Fourth, for EGMM, the marginalization method performed consistently and surprisingly well, so we recommend it as the best method to use. Finally, we observe that mice imputation worked quite well across all three anomaly detection methods, so we recommend it as the first method to try with any new anomaly detectors. 

\section{Acknowledgments}
This material is based upon work supported by the National     Science Foundation under Grant No.~1514550.

\bibliographystyle{ACM-Reference-Format}
\bibliography{references}

\appendix
\section{Synthetic Data Generation Details}
{\bf Correlated:} To generate each point $x$ of the nominal distribution, we first sample a mean vector $\vec{\mu}$ uniformly uniformly along the multi-variate diagonal line segment that extends from $(-3,\ldots,-3)$ to $(+3,\ldots,+3)$. Then $x$ is sampled from $\mathcal{N}(\vec{\mu}, \Sigma)$, where $\Sigma = diag(c) + \rho - diag(c\rho)$ and $\rho\in \{0.4, 0.6, 0.8, 1.2\}$. Each value of $\rho$ was used 5 times to obtain a total of 20 data sets. Anomalies are generated in a similar way but using the diagonal line that goes from $b\vec{v} + (-3,\ldots,-3)$ to $b\vec{v} + (+3,\ldots,+3)$, where $\vec{v}$ is the vector $(+1,-1,+1,$ $-1, \ldots,+1,-1)$ and $b=2$. The vector $b\vec{v}$ offsets the anomaly points from the nominal points in an orthogonal direction.

{\bf Mixture:} Nominals are generated from $\frac{1}{3} \mathcal{N}(\mu_1,\Sigma_1)+\frac{1}{3} \mathcal{N}(\mu_2,\Sigma_2)+\frac{1}{3} \mathcal{N}(\mu_3,\Sigma_3)$ where $\mu_1 = (-3, -3, -3, \ldots, -3)$, $\mu_2 = (3, -3, 3, \ldots, -3)$, and $\mu_3 = (3, 3, 3, \ldots, 3)$. The covariance matrices are created using Cholesky the decomposition: $\Sigma_i = L_iL_i^T$, where $L_i$ is the Cholesky decomposition matrix of $\Sigma_i$. Each $L_i$ is defined as follows:  
$L_1 = L_3 = diag(cd) + \rho - diag(\rho,d)$ and $L_2= diag(d) + \rho - diag(c\rho,d)\}$, where $\rho\in \{0.4, 0.6, 0.8, 1.2\}$. Each value of $\rho$ was used 5 times to obtain a total of 20 data sets.  Anomalies are generated from a single Gaussian. The mean is constructed by first sampling its components from $Unif(-1,1)$ and then adding an offset of $(2, -2, 2, \ldots, 2)$. The covariance matrix is the identify matrix. This positions the anomalies to be near the center of the three nominal components.

\section{Graphs of All Synthetic Data Results}

\begin{figure}[b]
\centering
\includegraphics[width=\columnwidth]{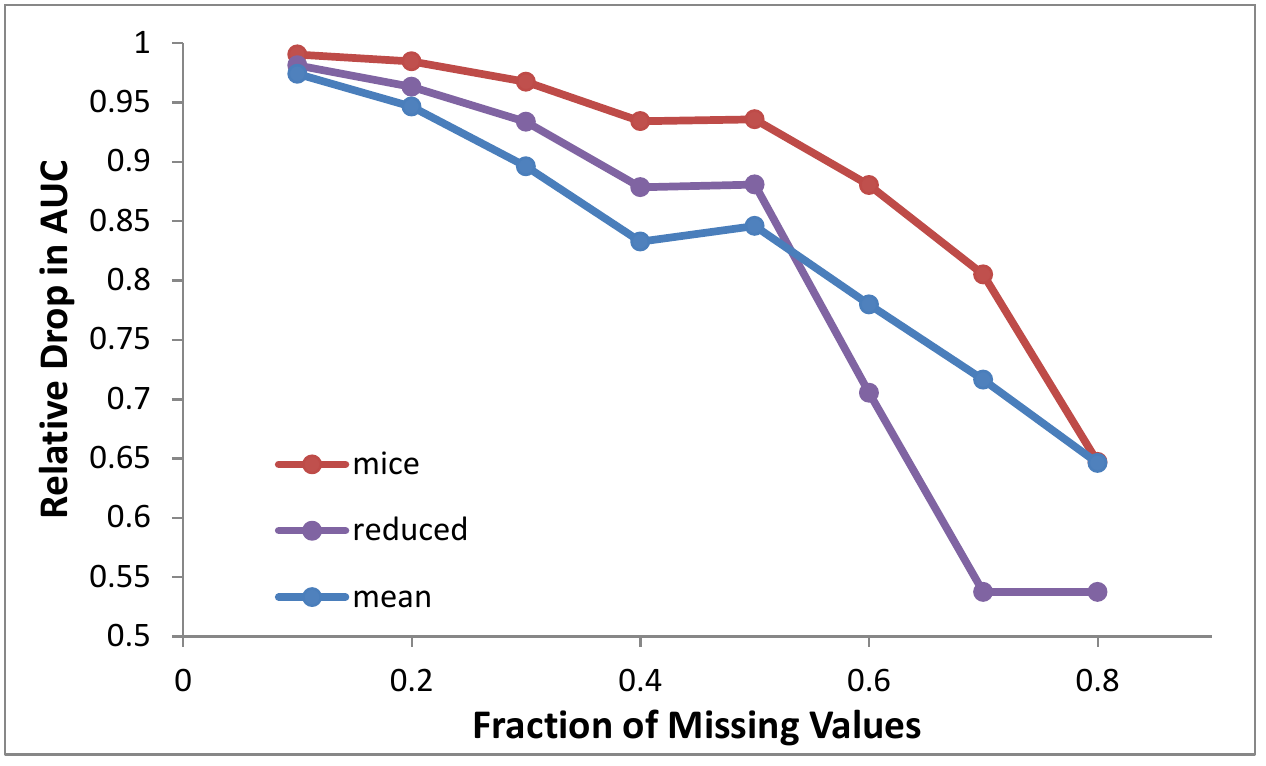}
\caption{Mean relative AUC of LODA as a function of the fraction of missing values, $\rho$, averaged across the four synthetic data set configurations.}
\label{fig:synthetic-mean-decay-loda}
\end{figure}

\begin{figure}[b]
\centering
\includegraphics[width=\columnwidth]{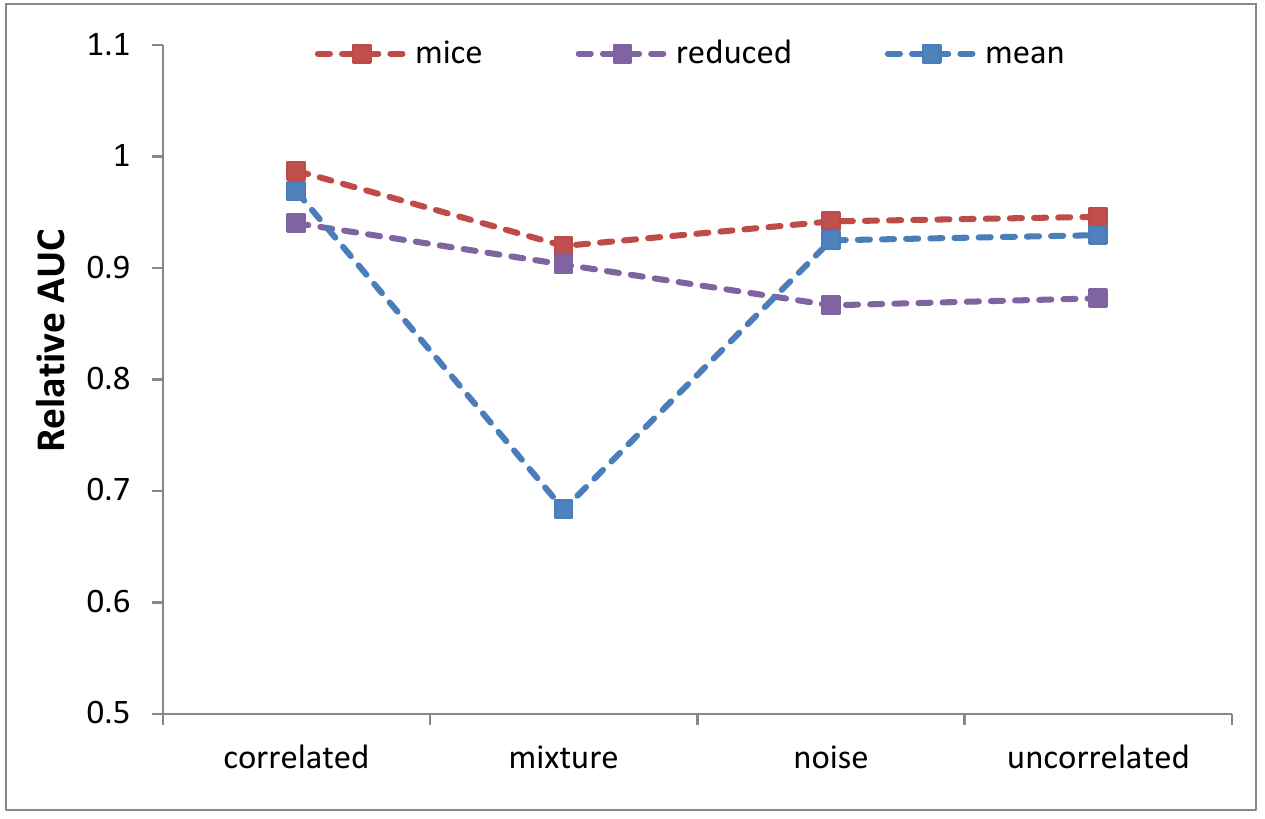}
\caption{Relative AUC of LODA on the synthetic data sets for $\rho=0.5$, in ascending order by mice imputation.}
\label{fig:rho=0.5-loda-synthetic}
\end{figure}

\begin{figure}[b]
\centering
\includegraphics[width=\columnwidth]{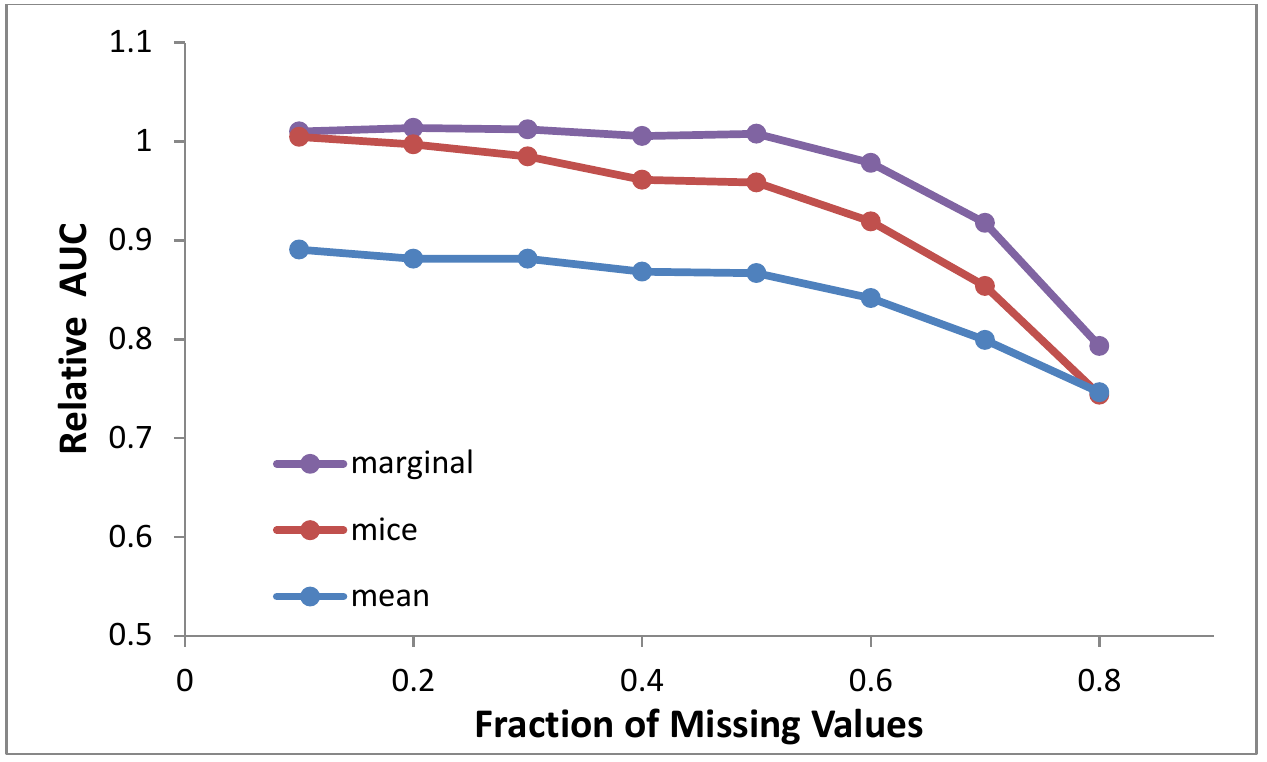}
\caption{Mean relative AUC of EGMM as a function of the fraction of missing values, $\rho$, averaged across the four synthetic data set configurations.}
\label{fig:synthetic-mean-decay-egmm}
\end{figure}

\begin{figure}[b]
\centering
\includegraphics[width=\columnwidth]{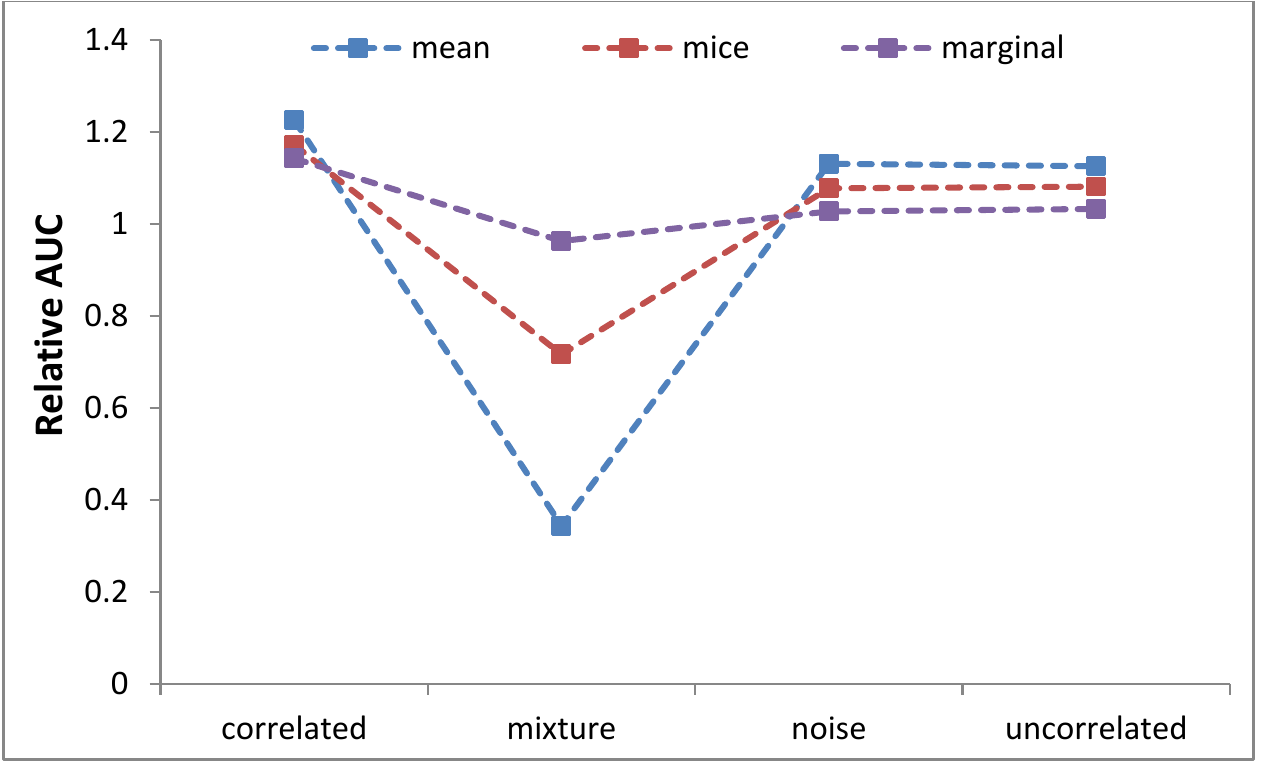}
\caption{Relative AUC of EGMM on the synthetic data sets for $\rho=0.5$, in ascending order by mice imputation.}
\label{fig:rho=0.5-egmm-synthetic}
\end{figure}

Figures~\ref{fig:synthetic-mean-decay-loda} and \ref{fig:rho=0.5-loda-synthetic} show the behavior of LODA on the synthetic data sets. We observe that mice is much better than the other methods on all four configurations and that it always beats LODA's own reduced method. Figures~\ref{fig:synthetic-mean-decay-egmm} and \ref{fig:rho=0.5-egmm-synthetic} show the behavior of EGMM on the synthetic data sets. We observe that the marginal method generally gives the best performance across all configurations. However, this is primarily due to its excellent performance on the Mixture configuration, as mean imputation works slightly better on the other three configurations. 

\end{document}